\documentclass[twocolumn,10pt]{adon-generic-paper-2col}
\usepackage{microtype}
\usepackage[utf8]{inputenc}

\usepackage{times}      		
\usepackage{mathptmx}  		  
\usepackage{parskip}  			
\usepackage{booktabs}				
\usepackage{colortbl}				
\usepackage{placeins}				
\usepackage{float}
\usepackage[square,numbers]{natbib}
\title{Fully Convolutional Network for Melanoma Diagnostics}

\author[1,*]{Adon Phillips}
\author[2]{Iris Y.H. Teo}
\author[1]{Jochen Lang}
\affil[1]{School of Electrical Engineering and Computer Science, University of Ottawa,  Ottawa, K1N 6N5, Canada}
\affil[2]{Department of Pathology and Laboratory Medicine, University of Ottawa, Ottawa, K1H 8M5, Canada}
\affil[*]{aphil037@uottawa.ca}

\begin{abstract}
This work seeks to determine how modern machine learning techniques may be applied to the previously unexplored topic of melanoma diagnostics using digital pathology. We curated a new dataset of 50 patient cases of cutaneous melanoma in whole slide images (WSIs). We provide gold standard annotations for three tissue types (tumour, epidermis, and dermis) which are important for the prognostic measurements known as Breslow thickness and Clark level. Then, we devised a novel multi-stride fully convolutional network (FCN) architecture that outperformed other networks trained and evaluated using the same data according to standard metrics. Finally, we trained a model to detect and localize the target tissue types. When processing previously unseen cases, our model's output is qualitatively very similar to the gold standard. In addition to the standard metrics computed as a baseline for our approach, we asked three additional pathologists to measure the Breslow thickness on the network's output. Their responses were diagnostically equivalent to the ground truth measurements, and when removing cases where a measurement was not appropriate, inter-rater reliability (IRR) between the four pathologists was 75.0\%. Given the qualitative and quantitative results, it is possible to overcome the discriminative challenges of the skin and tumour anatomy for segmentation using modern machine learning techniques, though more work is required to improve the network’s performance on dermis segmentation. Further, we show that it is possible to achieve a level of accuracy required to manually perform the Breslow thickness measurement.
\end{abstract}


\begin{document}

\twocolumn[{%
  \begin{@twocolumnfalse}
    \maketitle
    \begin{abstract}
      \lipsum[1-2]
    \end{abstract}
  \end{@twocolumnfalse}
}]


\section{Introduction}
Cutaneous melanoma (CM, melanoma) is an aggressive form of skin cancer originating in melanocytes which are the cells responsible for the pigmentation of skin, hair and eyes. Melanoma develops as the result of DNA damage usually as a consequence of intense exposure to ultraviolet radiation. When a melanocyte suffers sufficient genetic damage it may begin to grow in an abnormal way and become a melanocytic tumour \citep{Kumar2014}. In the earliest stage of development the tumour will grow laterally along the epidermal layer of the skin. At this point complete surgical excision of the lesion is possible. However, if left undiagnosed the tumour may thicken and infiltrate deeper into the skin. If the tumour growth invades lymphatic vessels, cancer cells may separate from the primary tumour and travel to regional lymph nodes as metastatic disease. The tumour may also invade blood vessels causing cancer cells to be transported to distant locations resulting in metastatic cancer \citep{Abeloff2008}.

In the United States melanoma is the second most common cancer in men aged 20-39 and the fifth most common in men of any age. In women it is the third most common cancer through ages 20-39 and the sixth most common in women of all ages \citep{Society2017, melanoma_facts2017}. In places such as New Zealand and Australia, melanoma incidence is reaching epidemic levels especially in young people \citep{Glazer2017}. Fortunately, if diagnosed early, the five year survival rate is as high as 98\%. If the tumour has spread to regional nodes, five year survival is reduced to 62\% and then to only 18\% in the case of metastatic disease \citep{Society2017}. While melanoma represents 3\% of all skin cancers, it is responsible for 65\% of skin cancer deaths \citep{Orzan2015}. Clearly the importance of early detection and accurate diagnosis cannot be overstated.

The gold standard for the diagnosis of melanoma requires full-thickness excisional biopsy of suspected lesions followed by the histopathologic analysis of hematoxylin and eosin (H\&E) stained tissue sections \citep{Goodson2009}. Historically, and to this day this analysis is conducted under a light microscope for the vast majority of cases. However the recent availability of slide scanning technology, which is able to digitize prepared tissue specimens on glass slides into high resolution files has brought the medical specialty of pathology into the midst of a transition towards a digital workflow. As a consequence, this move to `Digital Pathology' (DP) has also afforded the computational image analysis community access to whole slide images (WSI) allowing the application of modern image processing techniques to the processing of such specimens \citep{Madabhushi2016}. This includes automated, computer-aided, and other diagnostic tools that have the ability to augment current clinical workflows in the hopes of improving patient care and outcomes.

\begin{figure*}[ht]
\centering
\includegraphics[scale=0.15]{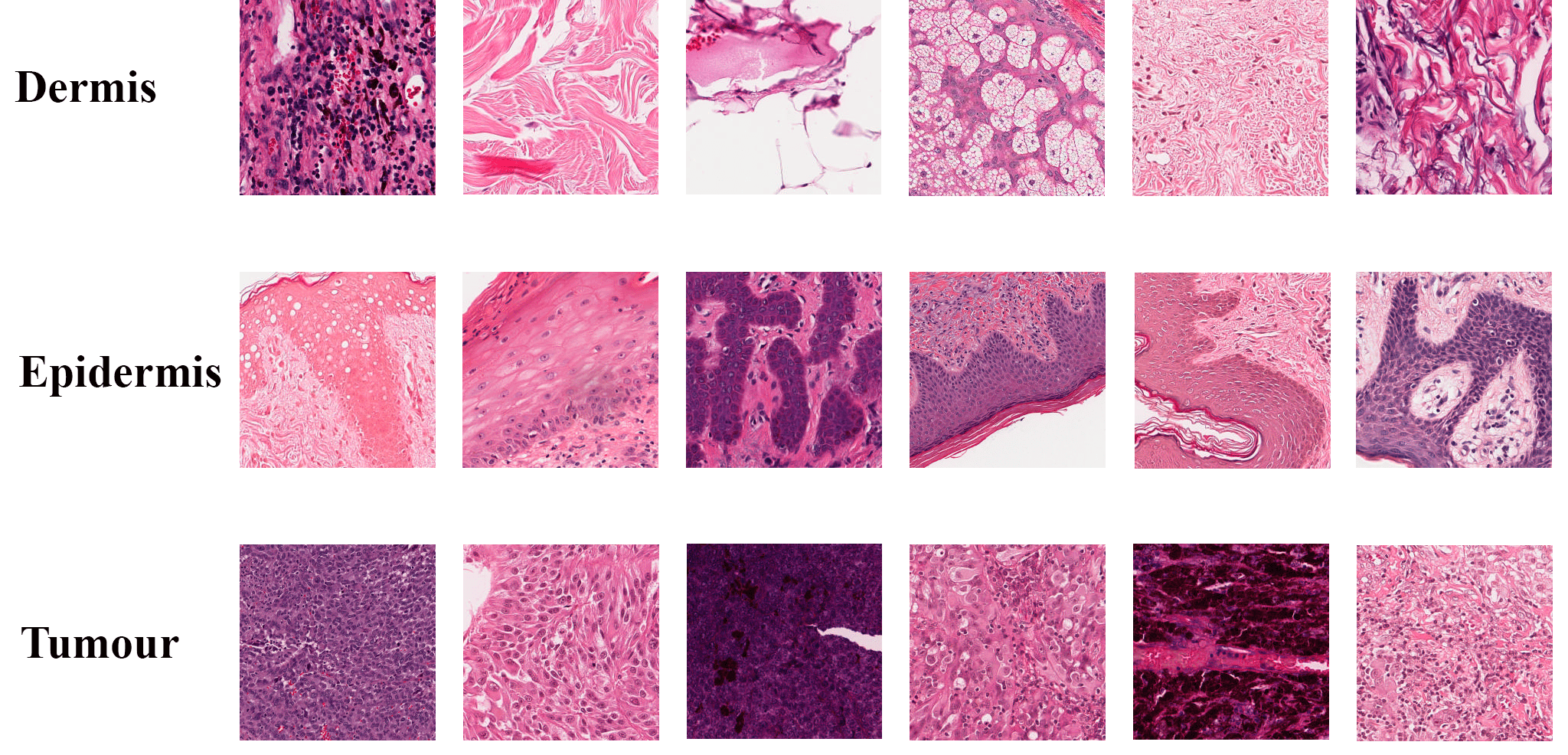}
\caption[Tissue variation examples by type]{A selection of patches to illustrate the variation between structures and staining for each tissue type in our data set.}
\label{fig:tissue_variation}
\end{figure*}

A major component of the diagnostic process is a pattern recognition task where the pathologist uses visual information combined with deep domain knowledge to identify complex histologic and cytologic features. In the case of cancer diagnosis, the variance in slide preparation, nuances of anatomy relating to qualitative and quantitative criteria of diagnosis, not to mention variability introduced by different slide scanners and associated software processing creates a feature space that is beyond the capacity of most handcrafted descriptors/extractors to account for. It has been shown that deep learning techniques have the potential to overcome such challenges, but in the case of melanoma diagnosis in WSIs which to our knowledge has not been explored, we will demonstrate that it is possible to accurately predict segmentation masks for multiple tissue types whose structures are vastly different in scale, and morphology with dramatic variations even between tissues of the same type as is the case in the pathology of the skin (see~\autoref{fig:tissue_variation}).

Furthermore, in addition to the diagnosis there is also a list of pertinent features for the prognostication and management of a given disease.  In the case of melanoma, the most important prognostic factor is the maximum tumour thickness, sometimes referred to as the Breslow thickness. According to the College of American Pathologists melanoma reporting protocol, it is considered throughout North America as the standard for reporting \citep{David2009}. While we do not know the segmentation accuracy required for such measurements in prognosis, we will show that it is possible to produce segmentation masks that are qualitatively, of sufficient accuracy to perform the Breslow thickness measurement.

We make the following major contributions in this work:
\begin{enumerate}
\item We curated a dataset of whole slide images containing cutaneous melanoma specimens, including full resolution annotations of tumour, epidermis, and dermis tissues by the expert dermatopathologist on our team.
\item We developed a neural network architecture specifically designed to incorporate features at multiple levels of granularity present in the images contained in our dataset for supervised learning.
\item We trained a model using our dataset and architecture to detect and localize structures important for melanoma diagnostics.
\item We assembled a panel of four pathologists to independently measure Breslow thickness on four cases from the test set and we computed standard segmentation metrics on model inference.
\end{enumerate}

\section{Related Work}


Much of the work in digital pathology involves the detection and segmentation of cellular structures such as nuclei, mitoses, individual cells and in our case epidermis and dermis layers of the skin. The motivation for identifying these structures relates to their role in various quantification methods as part of the diagnosis or staging of the disease. Here we briefly outline works in this area that are closely related to our work: \citet{Janowczyk2016a} developed a resolution adaptive deep hierarchical learning scheme for nucleus segmentation. \citet{kashif2016} combined hand-crafted features computed using the scattering transform for texture based features with a CNN to detect nuclei in cells which performs better than CNN alone. \citet{Turkki2016} created a method to detect tumour-infiltrating lymphocytes by classifying superpixels. Their CNN method outperformed texture based features and was equivalent to agreement between human pathologists. \citet{Veta2016} used a CNN based regression model to measure the mean nuclear area without additional steps in $4000\times4000$ pixel images. They converted their network to a fully convolutional network (FCN) for inference. \citet{DBLP:journals/corr/abs-1709-02554} used an encoder-decoder model to perform multi-class semantic segmentationa on breast tissue. Their method proposes dense connetions, where instead of a single connection between encoder and decoder, they connect a decoder to all other decoders at the same or lower level. They use a patch-based approach to train their model using their dataset of 58 whole slide images with patch size of $256\times256$ compared with our $512\times512$. They showed their approach was superior to traditional encoder-decoder networks and to an SVM approach. \citet{XU2018124} segment the dermis and epidermis regions in WSIs of skin based on thresholding the probability density function of the red channel, k-means and curve fitting using application specific image processing techniques as well as heuristics.

We investigated multiple approaches to class balancing for the dataset. In cases where the imbalance is severe it is common to adapt the loss function to include class weights. In their MRI based tumour segmentation network, \citet{Christ2017} added a weighting factor in the cross entropy loss function of their FCN and used an inverse class probability weighting scheme to deal with the class imbalance. \citet{Kamnitsas2017} partially allievitate class imabalance by sampling input segments of the source 3D brain MRI data such that training batches contain segments with a 50\% probability of being centered on a foreground or background voxel. Another common approach is to perform an undersampling of the majority class to bring the classes into alignment. This method can be problematic as there may be information lost that could be important for learning, though in cases where a sufficient number of samples are available it can be a workable technique \citep{Krawczyk2016}. In cases where data is scarce, \citet{Yen2009} proposed a cluster-based approach to selecting samples. The main idea is to divide the dataset into $k$ clusters and choose the clusters that will best represent the majority class according to criteria specific to the dataset at hand. \citet{Rahman2013} used a variation of Yen and Lee's approach to balance a challenging cardiovascular dataset. As an alternative to undersampling, especially in cases of limited data, oversampling the minority class may be appropriate. In this case the dataset distribution can be modified by duplicating training examples towards a dataset-appropriate metric \citep{Zhou2006}. The result is that the minority class will have a similar number of examples as compared with the other classes in the dataset. The derivation of the `duplicate' examples varies according to the problem.

%

The FCN style architecture on which we have based our network, was introduced by \citet{long2015} which has become the state-of-the-art approach for semantic segmentation in natural images \citep{Wang2017}. This approach trains the network end-to-end to obtain dense pixel-wise predictions. Inputs to FCNs can be images of arbitrary size, though we use patches due to the gigapixel scale of our source images  which well exceed current GPU memory capacity; this has the unfortunate effect of reducing the global information available especially in context of the overall slide. However, this patch-based approach is common when working with WSIs (\citet{HouSKGDS16}, \citet{Qaiser2017}, \citet{Janowczyk2016a}, \citet{DBLP:journals/corr/abs-1709-02554}).

Prior to the introduction of FCNs, classification networks were having great success in coarse inference, but the next step was dense, pixel-wise prediction models. In a general classification network the fully connected layers discard spatial information as the order of input pixels will not affect the output order. This reduction of context was detrimental to the performance of such approaches and would likely be detrimental to our task. Humans make extensive use of context such as relative size, relative location, probability of objects being in some `scenes' but not in others, when detecting objects and comprehending scenes \citep{Biederman1982}. This is also the case in semantic segmentation networks as shown by \citet{Mostajabi2015} and \citet{Liu2015}, who demonstrated great improvements on standard datasets by providing global and local context to their networks.

To retain the spatial information in FCNs, Long \etal replaced the fully connected layers in classification nets with convolutional layers. In doing so, the network computes a nonlinear filter down to the output map which contains the localized activations as predictions. At this point however, the predictions are in a down-sampled dimensional space of the input as a result of the convolution and pooling layers of the network. To up-sample the output map to the correct size, Long \etal use convolutional layers with a fractional input stride (combined with an activation function) to learn the up-sampling as a filter within the network (often called deconvolution) \citep{Zeiler2010}. The ability to control the input stride for the deconvolution allows the network to learn up-sampling filters of selectable granularity. We exploit this ability to capture multiple levels of granularity and fuse the results. Since Long \etals publication, FCNs have become the foundation for many segmentation architectures in convolutional neural networks regardless of application domain \citep{LITJENS201760}.

\section{Dataset and Preprocessing}

Our dataset includes 50 whole slide images (WSIs) containing hematoxylin and eosin (H\&E) stained, full-thickness excisions from 49 individuals. Source images were scanned at $40\times$ objective magnification using Aperio ScanScope slide scanners from nine different medical institutions. Annotations are provided at the equivalent resolution to the $40\times$ magnification source images. The overall dataset is divided into training, validation, and test sets at a ratio of $70{:}15{:}15$ percent. The data is grouped by patient to allow stratification by patient such that pixels from a given patient only appear in one of the training, validation, or test datasets. The dataset consists of a source WSI and its corresponding annotation image. Annotation images contain labels for tumour, epidermis and dermis tissue slides as required for tumour detection and staging. Annotations also include classes for background, and a catch-all label for structures that are not diagnostically important (NDI). Annotation images saved as five colour indexed PNG files. The source images for the dataset were obtained from the Human Skin Cutaneous Melanoma project (HSCM), which is part of The Cancer Genome Atlas (TCGA).


\subsection{Case Selection Criteria}

The one thousand melanoma slides were manually filtered at very low resolution to identify H\&E stained specimens containing epidermis and tumour cells. From this review 150 slides were identified for possible inclusion.

From the 150 candidate slides, inclusion was based on three main criteria:
\begin{enumerate}
\item To ensure more uniform quality of staining (to the degree possible) and to facilitate interpretation, material which appeared to be formalin-fixed paraffin-embedded tissue was selected over material with frozen section artifact (the dataset does not include any tissues from frozen sections).
\item The histology needed to be interpretable; tumour cells needed to be readily distinguishable from histiocytes, lymphocytes, and other stromal cells.
\item Slides with higher proportion of uninvolved epidermis or with slides with epidermis present were also favoured, due to the requirement for epidermis to be present for the Breslow thickness measurement, and the need to maximize the number of epidermis samples in the dataset.
\end{enumerate}

\subsection{Annotation strategy}
As most of the selected WSIs contained tumours that were mass forming, the external edge of the tumour was outlined. Large areas which were composed of inflammatory cells were not marked as tumour; however, if melanoma cells resided within the inflammatory population and could be definitively identified, these individual cells were marked. The blood vessels and stroma of the tumour were also included with the tumour, as were tumour infiltrating lymphocytes within the borders of the tumour. Necrotic areas, apoptotic cells, and any areas of cystic degeneration were also marked as tumour. At ulcer sites, only tumour cells were marked, and areas of fibrinopurulent exudate were not marked as tumour. Where they could be positively distinguished from mimickers, melanoma cells within the epidermis (in-situ melanoma) were also marked but were not actively sought out.

On some slides, tumour fragments within lymphatic or small thin-walled vessels were highlighted on a separate layer as lymphovascular invasion for possible inclusion in future work.

The epidermis was labeled from the basement membrane to the top of the granular layer (the outer edge of the epidermis). Markedly reactive epidermis and tangentially sectioned epidermis was included. The stratum corneum (the outermost layer of the skin) was not labelled. As the infundibular portion of the hair follicle derives from the epidermis, this was included in the epidermis markings. In spite of its morphologic similarity to the epidermis, the remainder of the hair follicle was included within the dermis layer as it represents a dermal structure. This decision requires further analysis to determine the impact on the prediction of the included tissue types.

The lower boundary of the dermal layer was  difficult to define, as it undulates and merges imperceptibly with the fibrous septae of the underlying subcutaneous adipose tissue and an arbitrary cutoff had to occasionally be employed.

Annotations were perfomed by anatomical pathologist with expertise in dermatopathology.

\subsection{Data Augmentation}

The thirty six patient cases included in the training set translates to over 62,000 image patches at $512 \times 512$ pixels or approximately 16.3 billion pixels. Based on other similar works using semantic segmentation \citep{Janowczyk2016, Cruz-Roa2014,lin2016,Menze2015}, this represents a sufficiently large collection of data for experimentation. However, as is the case with many medical imaging datasets, due to the natural distribution of the tissue types in our dataset, there exists a class imbalance between certain tissue types and between tissue types and the background.

\begin{figure}[ht]
\centering
\captionsetup{margin=0.4cm}
\includegraphics[width=\columnwidth]{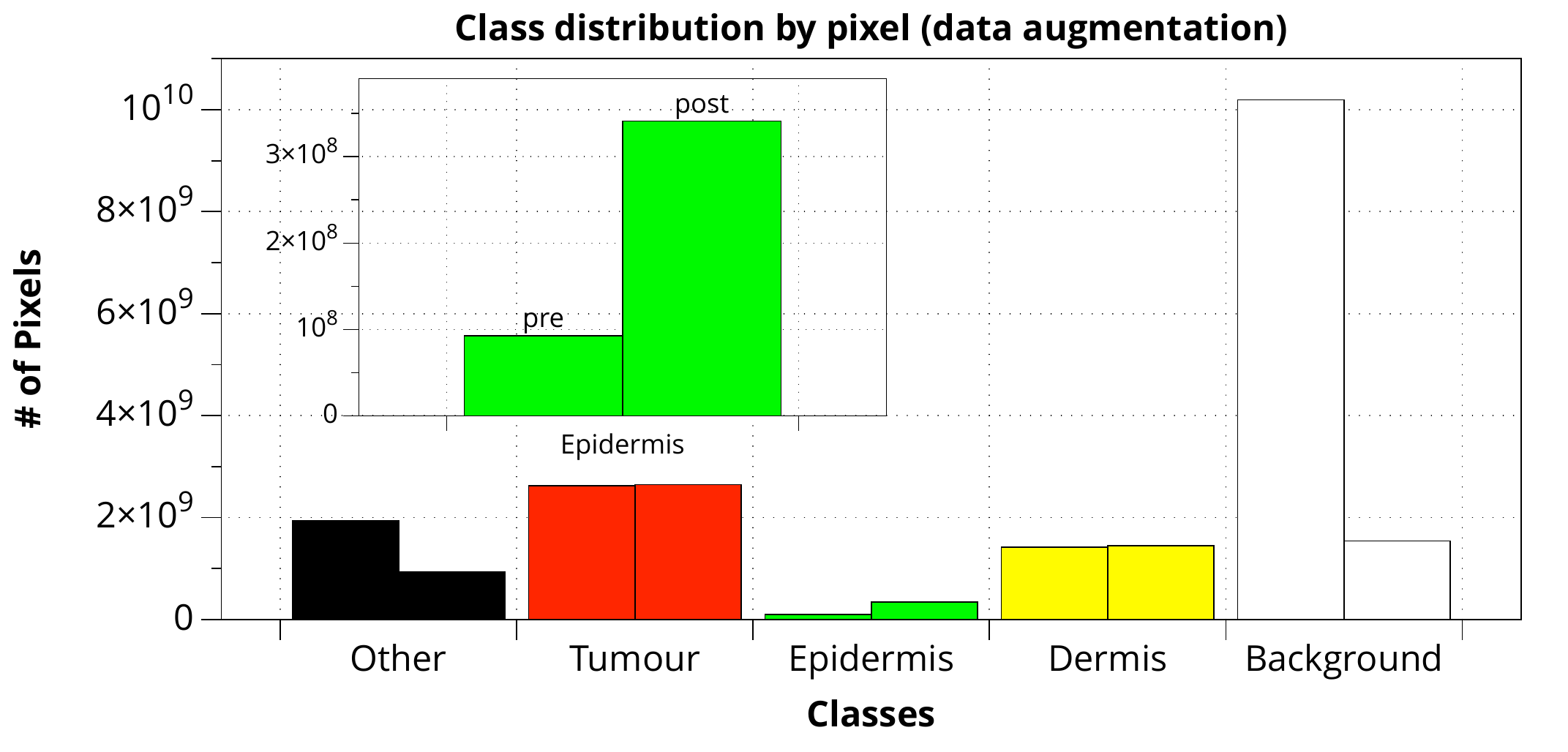}
\caption[Post augmentation histogram, stage 2]{Class distribution by the number of pixels per class following the described data augmentation strategy showing the addition of new epidermis only patches (minimal effect).}
\label{fig:da_post2}
\end{figure}

Our dataset suffers from two major problems. First, since WSIs consist of a tissue specimen placed in the center of the slide, and due to the background being visible through the tissue specimen, background pixels are the most common by far. Background pixels outnumber the second most common class (tumour) $4{:}1$. Most of these pixels are are in patches that only contain background pixels. Secondly tumour labelled pixels outnumber epidermis labelled pixels $28{:}1$ (see \autoref{fig:da_post2}). Initial experiments with the unmodified data showed a test network would converge with high accuracy by predicting the tumour class for all tissue types (severe overfitting). Furthermore, early experiments also showed there were insufficient pixels labelled in the epidermis class to train the network to recognize such regions accurately in the face of such a large imbalance.


To address these problems we have formulated a hybrid class balancing solution to effectively undersample and oversample the problematic classes as required:

\textbf{Undersampling}

As stated, most of the background pixels are found in patches containing only background, from areas around the periphery of the WSIs. We undersample by removing patches where background pixels are most common and make up more than $98\%$ of the target patch. This percentage retains patches that might contain locally important features. We also remove patches where background pixels are most common and the NDI class is the only other class present.

\textbf{Oversampling}

For the epidermis class which is the minority class, we oversample using basic data augmentation transformations to increase the number of epidermis containing patches with respect to patches of other classes. We parse the pixels of each patch containing the annotations and store the list of patches that meet the following criteria:
\begin{itemize}
\item All pixels are epidermis
\item Most pixels are background, second most pixels are epidermis
\item Most pixels are background, second most pixels are NDI, third most pixels are epidermis
\item Most pixels are NDI, second most pixels are epidermis
\item Most pixels are NDI, second most pixels are NDI, third most pixels are epidermis
\end{itemize}

These annotation patches and the corresponding feature patches are then augmented with the following transformations: Flip left to right, flip top to bottom, rotate $90^{\circ}$, rotate $270^{\circ}$.

The result of the proposed balancing scheme is a generally balanced dataset with a modest but tractable imbalance remaining between the tumour, and epidermis classes. This imbalance was reduced from $28{:}1$ to $7.75{:}1$ (see \autoref{fig:da_post2}). It might seem to make sense to undersample the tumour class to further reduce the imbalance, though after experimentation we found that the epidermis was receiving accurate segmentations compared to pre-balancing tests. Furthermore the anatomical variance of tumour cells and regions across patients and even individual slides is significantly greater than that of the epidermal cells (see \autoref{fig:tissue_variation} for an illustration of in-class variation). Therefore we thought it advantageous to retain as many tumour examples as possible.



\section{Network}

The network architecture we present is the result of a series of experiments to adapt a previous fully convolutional network (FCN) design by \citet{long2015} for our dataset. Our experiments sought to maximize the simple heuristic of increasing global context for local predictions while constraining GPU memory requirements, network size, and computational complexity.


The proposed network uses an FCN architecture with a structure based on the work of \citet{long2015} which itself is based on VGG16 \citep{Simonyan2014a} and was converted to an FCN by Long \etal The base structure uses a skip net which combines coarse predictions from deep layers with fine predictions from shallow layers to provide improved overall granularity in the output segmentation. The FCN-16s for example fuses predictions from the top layer of the stack with predictions from the fourth pooling layer at stride 16. The result is upsampled in a deconvolutional layer to the original input size. The fusion operation is implemented as an element-wise summation of scores from respective streams.

Our work explores modifications to the base network structure (FCN-16s) to improve the range of information available for dense prediction. The approach aims to include features at multiple levels of granularity via deconvolution using multiple pixel strides. The resulting scores can be fused and a weight assigned to each stride. In the standard FCN-16s style network, three levels of output granularity can be obtained based on three architectural variants in three discrete networks. These three network configurations range from a single stream with a 32 pixel stride (FCN-32s) to three streams fused into an 8 pixel stride (FCN-8s). In FCN-16s, scores from each stream are fused in a step-wise fashion so that intermediate scores from stream 1 and stream 2 are combined and upsampled using deconvolution, then that result is fused before being upsampled with a stride of 16 resulting in the final output score. This combination of layers facilitates local predictions while retaining some global context.

\begin{figure}[ht]
\centering
\includegraphics[scale=0.50]{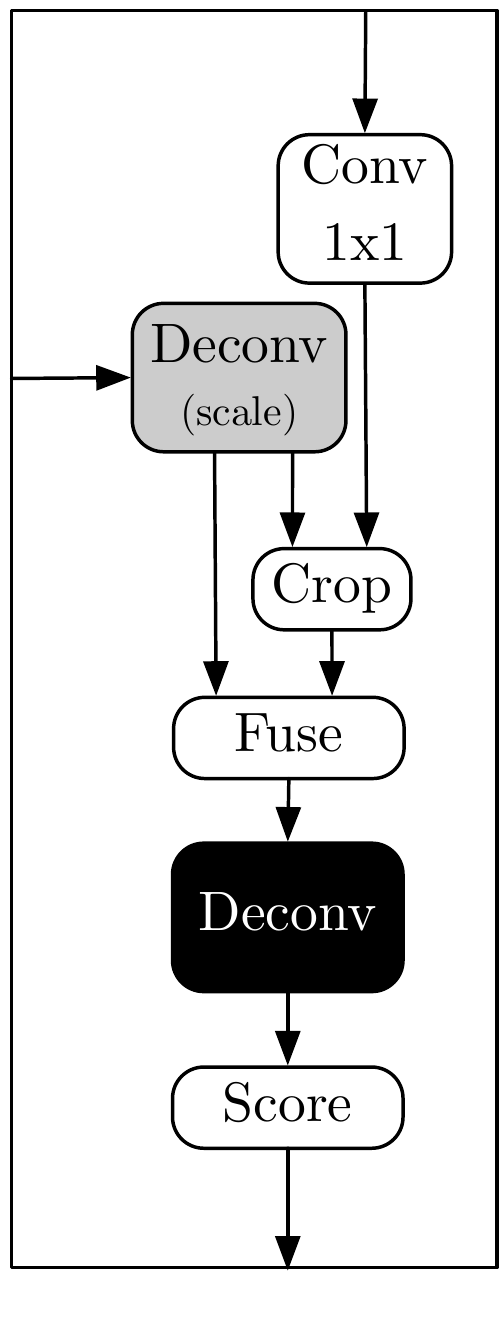}
\caption[Deconvolutional block.]{A representation of the deconvolutional block. The top input is the output from a previous convolutional layer, usually via skip connection from a more shallow postion in the network. The block also takes input from the previous deconvolutional block prior to deconvolution which contains the weights from the deepest convolutional layer in the network. The final score layer crops to the original input dimensions.}
\label{fig:ms-zoom}
\end{figure}

While the FCN of Long \etal includes information from multiple streams the final output granularity is fixed. In our experiments we found that configuring the network for fine granularity would perform poorly on coarse features of the input, negating the improved segmentation on features with finer details. To improve these results for our segmentation problem, we perform prediction based on varying levels of granularity that is robust to  factors like boundary contour complexity, feature size and patch level homogeneity. To incorporate information at multiple levels of prediction granularity we use a deconvolutional block (see \autoref{fig:ms-zoom}) to produce scores for each of the desired deconvolutional strides.

\begin{figure}[ht]
\centering
\includegraphics[scale=0.55]{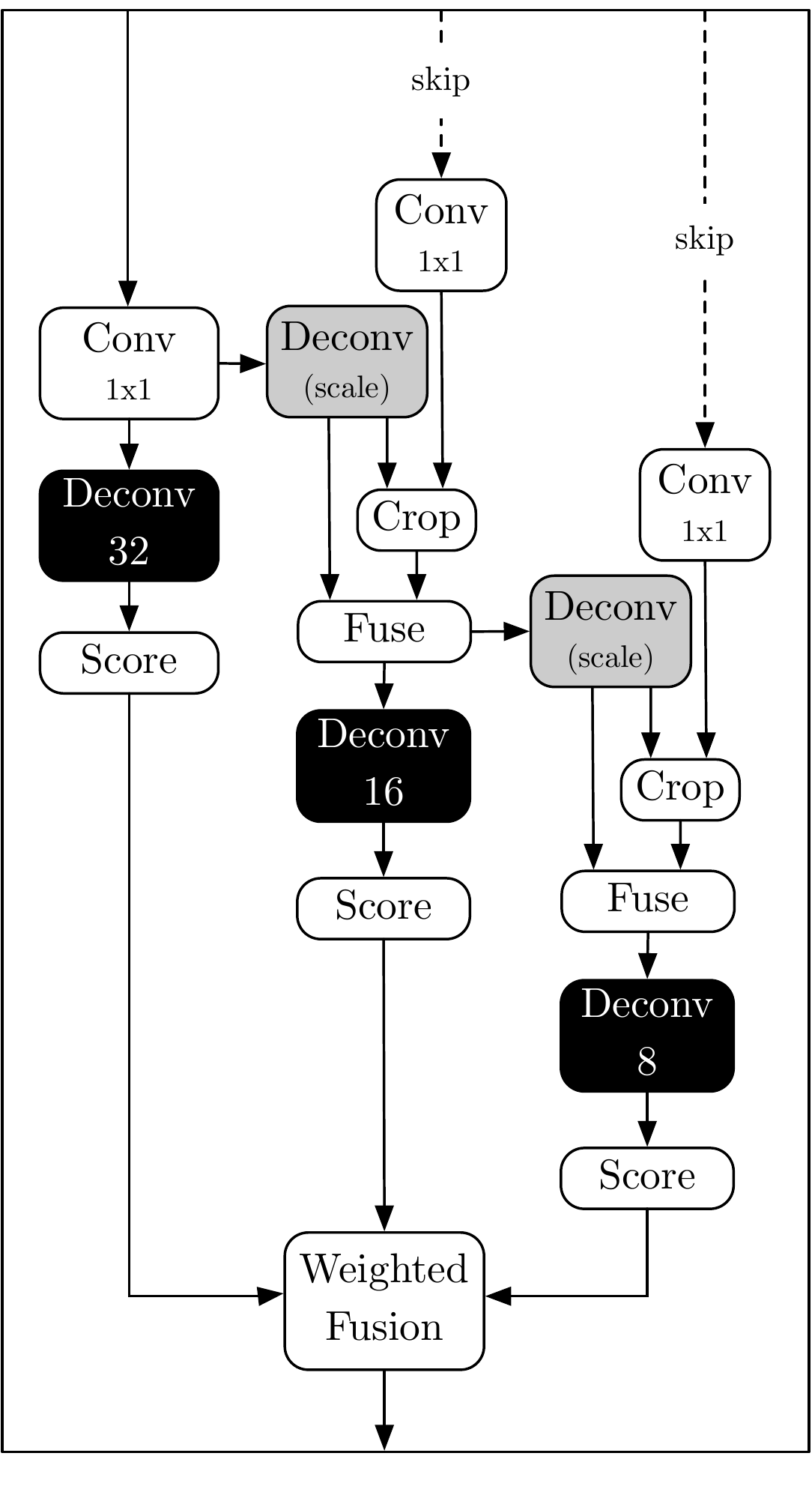}
\caption[Multi-stride fusion diagram]{A graphical overview of multiple deconvolutional blocks linked together. Scores from each deconvolution are combined using element-wise summation. This configuration represents that used to train our model. }
\label{fig:ms}
\end{figure}

We use Long \etal style deconvolutions that reverses the forward and backward passes with a fractional stride. Skip connections are added from progressively lower convolutional stacks to preserve global structure since finer strides see fewer pixels. Using the deconvolutional block, we compute three levels of granularity at three independent pixel strides (32,16,8). We then combine these scores using a weighted, element-wise summation, where per-stride weights are hyper-parameters for our model (see \autoref{fig:ms}) with coefficients of 0.5, 0.7, and 0.9 respectively.\\

\subsection{Training and Inference} \label{train-infer}
Experimental models were trained over 20 epochs on the same dataset. For additional training to produce the final results presented in \autoref{train-infer}, we initialized the network using pretrained weights from the initial 20 epoch training. From there we trained an additional 20 epoch round on the final training set.


Before training and inference, we computed the mean image from the respective datasets then normalized the inputs by subtracting the dataset mean image from each. This process was intended to shift the inputs so that the average was closer to zero. This has been shown to improve convergence times \citep{LeCun2012}.


All experiments and subsequent results were conducted using a single NVIDIA GTX 1080ti GPU. Models were trained using the Caffe open source deep learning framework \citep{jia2014caffe}. We used the `Lightning Memory-Mapped Database' (LMDB) to store dataset images. We also used NVIDIA DIGITS for some training experiments and visualizations.

\textbf{Hyperparameters}

We trained our networks using standard stochastic gradient descent. For most experiments we used a learning rate of $1 \times 10^{-4}$ with a sigmoid decay pattern to $1 \times 10^{-5}$. Experimental networks were trained on a subset of available patches (16,000 versus 26,000 in the full training set). To reduce overfitting we implemented a split dropout strategy using a rate of $0.9$ at the sixth fully convolutional layer (fc6) and $0.75$ at the deepest convolutional layer (fc7).

We used a minibatch of 2 across all networks. While \citet{long2015}, found a minibatch size of 1 to be optimal, we found no difference between size 1 and 2 in our approach.

\textbf{Transfer learning}

Notwithstanding the significant difference in domain specific features between the PascalVOC dataset and our medical images, transfer learning was essential to improving our training times and results. We initialized weights for learnable layers using publicly available (\url{https://github.com/shelhamer/fcn.berkeleyvision.org}) models\\ (FCN-32s, FCN-16s, FCN-8s) and fine-tuned. We used net-surgery to adapt existing models to our architecture when using multiple streams and layers are duplicated but names must change. These pretrained model architectures generally match our own up to the deepest convolutional layer. For our multi-stride networks we initialized weights using multiple pretrained networks trained at each requisite stride for the given deconvolutional block of our network. Non-matching weights were learned from random initializations.

\textbf{Loss function}





To compute training loss we used the softmax:
\begin{align}\label{eq:softmax}
  p(y^{(i)} = k\vert x^{(i)};\theta) = \frac{e^{\theta_k^T x^{(i)}}}{\sum_{k=1}^{K}e^{\theta_{k^\prime}^T x^{(i)}}},
\end{align}

with multinomial loss:
\begin{align}\label{eq:multinomial}
J(\theta) = \sum_{i=1}^{m}\sum_{k=1}^{K}1\{y^{(i)}=k\} \log p(y^{(i)} = k\vert x^{(i)};\theta),
\end{align}
where m is the sample and K is the class.

\subsection{Evaluation Metrics}\label{metrics}

In order to evaluate the performance of our proposed architecture on our curated dataset we have selected a set of commonly used metrics designed to measure the performance of semantic segmentation inference results \citep{Thoma2016}. We compute: Pixel accuracy, Mean Pixel Accuracy (mPA), Mean Intersection Over Union (mIoU) and Frequency Weighted Intersection Over Union (fwIoU). Inspired by \citet{zhou2017scene}, we compute a `Final Score' as the definitive metric for our evaluation. We use the mean pixel accuracy (mPA):
\begin{align}
\text{mPA} = \frac{1}{n_{cl}} \frac{\sum_i n_{ii}}{t_i},
\end{align}
and the mean intersection over union (mIoU):
\begin{align}
\text{mIoU} = \frac{1}{n_{cl}}\sum_i \frac{n_{ii}}{(t_i + \sum_j n_{ji} - n_{ii})}
\end{align}
and the following for the final score:
\begin{align}
\text{score} = \frac{\text{mPA}+\text{mIoU}}{2} 
\end{align}

\subsubsection{Breslow Measurement}

Pathology reports not only contain a diagnosis but also list pertinent features for the prognostication and management of a given disease. In cutaneous melanoma, the most important prognostic factor is the maximum tumour thickness, sometimes referred to as the Breslow thickness. According to the College of American Pathologists melanoma reporting protocol, considered throughout North America as the standard for reporting \citep{David2009}, this is typically measured at a right angle to the adjacent normal skin. Where the epidermis is intact, the  superficial point of reference is the upper edge of the granular layer of the epidermis; when the epidermis is lost, i.e., ulcerated, the base of the ulcer is used as the upper point of reference. The deepest point of the main mass of the tumour is taken as the deep point of reference.

To establish a quantitative accuracy measure for the qualitative inference results with respect to the Breslow thickness measurement, we asked four pathologists (including the original annotator) to mark the Breslow measurement on the inference output images using a small set of results where the measurement was diagnostically relevant (i.e. where tumour is primary site). After performing the annotations, the pathologists jointly reviewed the original WSIs corresponding to the inference results on which the Breslow measurement was marked to establish the ground truth for the Breslow thickness. Based on this ground truth, we calculate inter-rater agreement using Randolph's free-marginal multirater kappa method \citep{randolph}.


\section{Results and  Discussion} \label{res-disc}
\subsection{Qualitative Evaluation}
\begin{figure*}[!t]
\centering
\includegraphics[scale=.20]{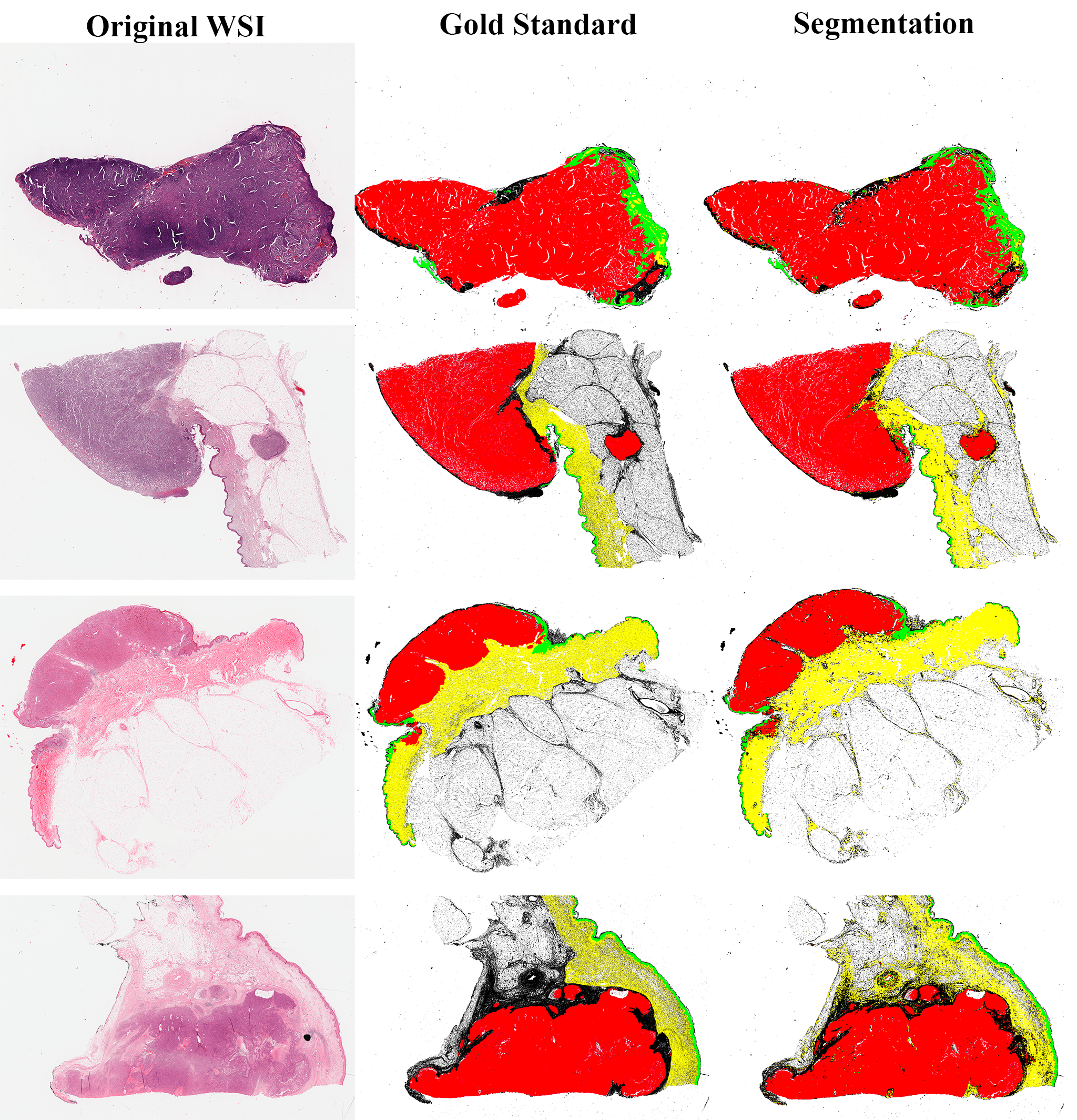} 
\caption[Whole slide image segmentation results]{Segmentation results from our model compared with the pathologist’s gold standard annotations.
Tumour shown in red, epidermis in green, dermis in yellow, other structures in black. The WSIs shown contain an average of 1300 patches.}
\label{fig:wsi_compare}
\end{figure*}

Using the  model trained on our dataset, using the process described in \autoref{train-infer},  \autoref{fig:wsi_compare} shows a strong correspondence between the gold standard and the inference predictions. We observe that the model is able to differentiate between tissue types and between highly varied morphology of the tissues. However the model struggles to infer the true depth of the dermis as the depth of the dermis is demarcated by the start of the adipose layer, but adipose cells are surrounded by collagen and other cells that make up the dermis.

\begin{figure*}[ht]
\centering
\includegraphics[scale=.11]{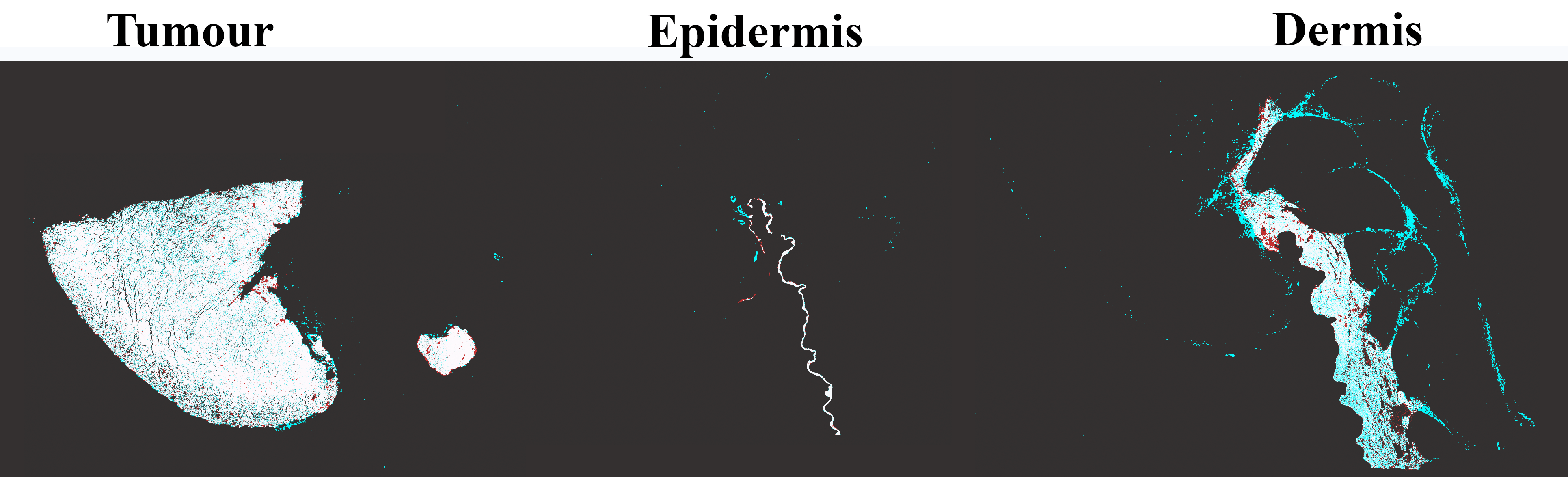} 
\caption{Visualization of segmentation errors by pixel. White represents true positive, teal represents false positive and red represents false negative by respective tissue type. The figure highlights the model's difficulty finding the boundary between the dermis and deeper tissues within the skin.}
\label{fig:wsi_accuracy}
\end{figure*}


Errors in tumour and epidermis tissue are largely found around patch boundary regions which do not affect practical accuracy (see \autoref{fig:wsi_accuracy}), but can lead to error accumulation. Instead of using non-overlapping patches, employing an overlapping strategy could ameliorate the observed border artifact. However this would be computationally expensive. Less common, serious errors could be addressed by increasing the size of the dataset, tuning the FCN and improving normalization.

\subsection{Quantitative Evaluation}
We have computed a set of standard metrics described in \autoref{tab:std_metrics} for each WSI in our test set as processed by our trained model. These metrics show high accuracy especially for pixel accuracy, which can be partially attributed to the performance in segmenting the background. However, the other metrics together support our qualitative observations on the quality of the segmentation, and the ability to use the resulting segmentation mask to measure the Breslow thickness.

\begin{table}[htbp]
  \centering
	\captionsetup{margin=0.4cm}
  \caption{Calculation of our selected standard metrics for inference results. The results were obtained from the test set consisting of seven whole slide images of various sizes and class composition trained over 40 epochs. See \autoref{metrics} for a description of the metrics. Time is measured from the point each patch enters the network until the output is calculated. The final time is the sum of this measure for all WSIs in the test set.}
		\resizebox{\columnwidth}{!}{
    \begin{tabular}{rrrrrrr}
    \toprule
    \multicolumn{1}{l}{\textbf{PA}} & \multicolumn{1}{l}{\textbf{mPA}} & \multicolumn{1}{l}{\textbf{mIoU}} & \multicolumn{1}{l}{\textbf{fwIoU}} & \multicolumn{1}{l}{\textbf{Score}} & \multicolumn{1}{l}{\textbf{Time (s)}} \\
    \midrule
    {0.8922} & {0.6307} & {0.5839} & {0.8650} & {0.6073} & {133.7284} \\
    \bottomrule
    \end{tabular} }%
  \label{tab:std_metrics}%
\end{table}%

\subsection{Breslow Measurement}
The results of a panel of four pathologists measuring Breslow on the output were diagnostically equivalent. When removing cases where measurements were not appropriate, inter-rater agreement was 75.0\% with a fixed marginal kappa of 0.5. This result supports the claim that the tumour and epidermis segmentation masks are sufficiently accurate to manually measure the Breslow thickness and thus it would be possible to use our approach to estimate the Breslow thickness automatically as part of an automated diagnostic workflow.

\subsection{Network Comparison}
We trained three distinct networks, including our multi-stride network and two additional, previously presented networks for comparison that have been used for similar tasks in digital pathology (see \autoref{tab:net_summary}). The purpose of these experiments was to evaluate our network's design against a competent baseline network before running the final inferenece results.

We trained these networks using the training set with a minibatch size of 2. We trained each for 20 epochs (approximately 167,000 iterations). All networks were fine-tuned using publicly available pretrained weights. We attempted training from scratch using FCN-16s with limited success and opted for fine-tuning as a time saving and performance improving measure. We also experimented with multi-scale information as a second branch with patches scaled to 20 $\times$ objective resolution. When combined with the multiple stride information the additional context availble in the lower resolution inputs afforded minimal improvements, but doubled the computation time.

\begin{table}[htbp]
  \centering
	\captionsetup{margin=0.4cm}
  \caption[Performance metrics summary for each network]{A summary of the network inference results showing the mean values computed across each of the test slides for each respective network. Bold text represents the best result for each metric. See \autoref{tab:std_metrics} for a description of the metrics.}
	\resizebox{\columnwidth}{!}{
    \begin{tabular}{lrrrrrr}
    \toprule
    \textbf{Network} & \multicolumn{1}{l}{\textbf{PA}} & \multicolumn{1}{l}{\textbf{mPA}} & \multicolumn{1}{l}{\textbf{mIoU}} & \multicolumn{1}{l}{\textbf{fwIoU}} & \multicolumn{1}{l}{\textbf{Score}} & \multicolumn{1}{l}{\textbf{Time (s)}} \\
    \midrule
    AlexNet & 0.8904 & 0.5300 & 0.4913 & 0.8495 & 0.5106 & \textbf{51.3423} \\
    FCN-16s & 0.8775 & 0.5513 & 0.5068 & 0.8433 & 0.5291 & 139.5002 \\
    Our Multi-Stride & \textbf{0.8910} & \textbf{0.6240} & \textbf{0.5781} & \textbf{0.8634} & \textbf{0.6011} & 134.0652 \\
    \end{tabular} }%
  \label{tab:net_summary}%
\end{table}%

\begin{figure}[ht]
\centering
\captionsetup{margin=0.3cm}

\includegraphics[scale=.2]{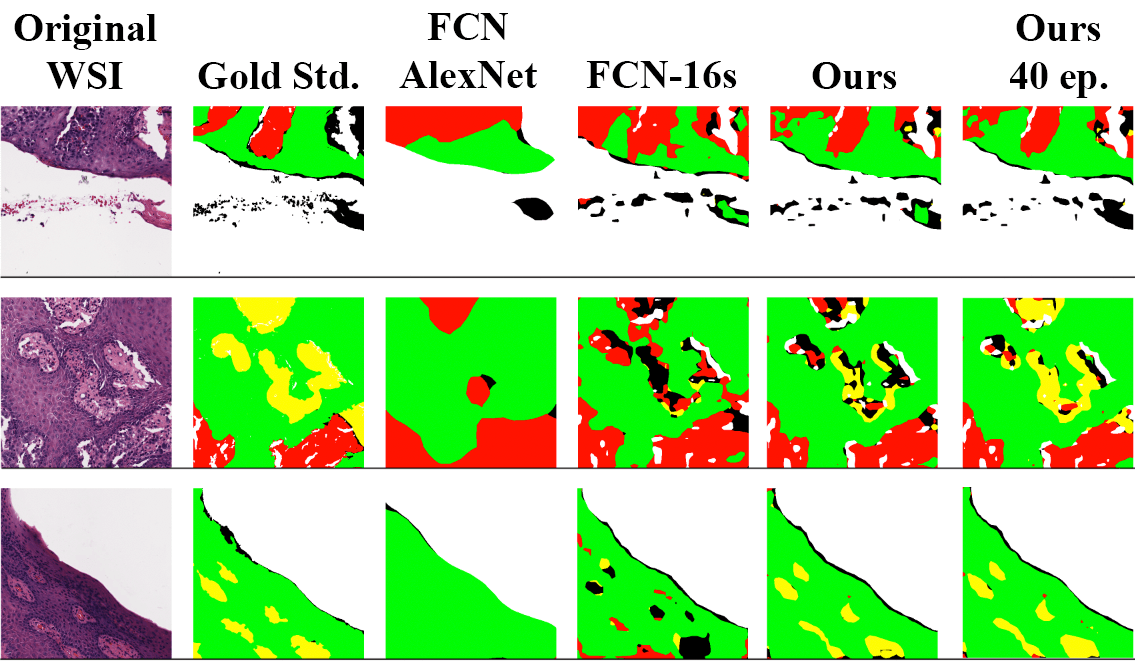} 
\caption{Example discriminatory diferences at the patch level. Three randomly selected patches compared by network. Each network was trained on the same data over 20 epochs. We include patches processed by our model trained over 40 epochs for reference. Note the yellow dermal regions.}
\label{fig:wsi_accuracy_patch}
\end{figure}

The results show our network outperforms the FCN-16s baseline network by 8\%. Qualitatively, the multi-stride information adds disciminatory power in cases where tissues are difficult to discriminate or contours in features are more complex in shape (see \autoref{fig:wsi_accuracy_patch} for patch comparison).


\section{Conclusions}
We have demonstrated a method capable of detecting and localizing melanoma tumours and other tissue structures important for prognostic measurements using a custom fully convolutional network on a dataset that we have curated.

Given the qualitative and quantitative results it is clearly possible to overcome the discriminative challenges of the skin and tumour anatomy for segmentation using modern machine learning techniques. Further we have shown it is possible to approach a level of accuracy to allow manual measurement of the Breslow thickess. More work is required to improve the network’s performance on dermis segmentation.

This work is a starting point for potential future applications in diagnostics workflows or within QA/QC workflows for retrospective review of past cases looking for diagnostic discrepancies.


\FloatBarrier


\end{document}